\documentclass[conference]{IEEEtran}
\IEEEoverridecommandlockouts
\usepackage{cite}
\usepackage{amsmath,amssymb,amsfonts}
\usepackage{algorithmic}
\usepackage{graphicx}
\usepackage{textcomp}
\usepackage{xcolor}
\usepackage{cleveref}
\usepackage{multirow}
\usepackage{url}
\usepackage{booktabs}   
\newcommand{\ours}{RoRA}
\def\BibTeX{{\rm B\kern-.05em{\sc i\kern-.025em b}\kern-.08em
    T\kern-.1667em\lower.7ex\hbox{E}\kern-.125emX}}

\begin{document}               
\title{RoRA: Efficient Fine-Tuning of LLM with Reliability Optimization for Rank Adaptation 
%{\footnotesize \textsuperscript{*}Note: Sub-titles are not captured for https://ieeexplore.ieee.org  and
}  
\author{
Jun Liu$^{1,2}$, Zhenglun Kong$^{1}$, Peiyan Dong$^{3}$, Changdi Yang$^{1}$, Xuan Shen$^{1}$, Pu Zhao$^{1}$, Hao Tang$^{2}$, Geng Yuan$^{4}$, Wei Niu$^{4}$,\\
Wenbin Zhang$^{5}$, Xue Lin$^{1}$, Dong Huang{$^{2,*}$}\thanks{* Corresponding authors.}, Yanzhi Wang$^{1,*}$ \\
\\
\textit{$^1$Northeastern University, Boston, USA} \quad
\textit{$^2$Carnegie Mellon University, Pittsburgh, USA} \\
% \textit{$^3$Swiss Federal Institute of Technology Zurich, Zurich, Swiss} \\
\textit{$^3$Massachusetts Institute of Technology, Boston, USA} \quad
\textit{$^4$University of Georgia, Athens, USA} \\
\textit{$^5$Florida International University, Miami, USA}
}

\maketitle

\begin{abstract}

Fine-tuning helps large language models (LLM) recover degraded information and enhance task performance.
Although Low-Rank Adaptation (LoRA) is widely used and effective for fine-tuning, we have observed that its scaling factor can limit or even reduce performance as the rank size increases. To address this issue, we propose RoRA (Rank-adaptive Reliability Optimization), a simple yet effective method for optimizing LoRA's scaling factor. By replacing $\alpha/r$ with $\alpha/\sqrt{r}$, RoRA ensures improved performance as rank size increases. Moreover, RoRA enhances low-rank adaptation in fine-tuning uncompressed models and excels in the more challenging task of accuracy recovery when fine-tuning pruned models.
Extensive experiments demonstrate the effectiveness of RoRA in fine-tuning both uncompressed and pruned models. RoRA surpasses the state-of-the-art (SOTA) in average accuracy and robustness on LLaMA-7B/13B, LLaMA2-7B, and LLaMA3-8B, specifically outperforming LoRA and DoRA by 6.5\% and 2.9\% on LLaMA-7B, respectively. In pruned model fine-tuning, RoRA shows significant advantages; for SHEARED-LLAMA-1.3, a LLaMA-7B with 81.4\% pruning, RoRA achieves 5.7\% higher average accuracy than LoRA and 3.9\% higher than DoRA.

\end{abstract}


\section{Introduction}
Large language models (LLMs) are typically trained on broad datasets during pretraining, which enables the model to develop general language understanding capabilities. Fine-tuning allows the model to perform better on specific tasks or domains. For example, fine-tuning can help the model better handle text in specialized areas such as healthcare or law. Fine-tuning helps reduce biases and generate more relevant and natural text. Moreover, Large-scale deep learning models
~\cite{together2023redpajama},\cite{puma3b},\cite{Mamba-GPT-3b-v2},\cite{openalpaca},\cite{xia2023sheared},\cite{zhang2022advancing},\cite{li2022pruning},\cite{yang2023pruning}
%~\cite{together2023redpajama,puma3b,Mamba-GPT-3b-v2,openalpaca},
which often comprise billions or even hundreds of billions of parameters, face limitations in deployment on resource-constrained devices~\cite{yuan2022you},~\cite{yuan2022layer}, \cite{yuan2021mest},~\cite{li2024waxing},~\cite{zhan2024exploring},~\cite{zhao-etal-2024-pruning}, such as mobile phones. Pruning techniques are commonly applied to address these challenges by reducing model size and computational overhead while maintaining performance~\cite{yuan2021forms},\cite{yuan2021tinyadc},\cite{gong2022automatic}.
Parameter-Efficient Fine-Tuning (PEFT)~\cite{peft} has also gained prominence as a strategy to reduce the high computational cost of full model fine-tuning. This method allows for efficient task-specific~\cite{liu2024tsla},~\cite{liu2023scalable} fine-tuning of large language models (LLMs) without the need to retrain all parameters.

\begin{figure}[tbp]
\centerline{\includegraphics[width=0.85\linewidth]{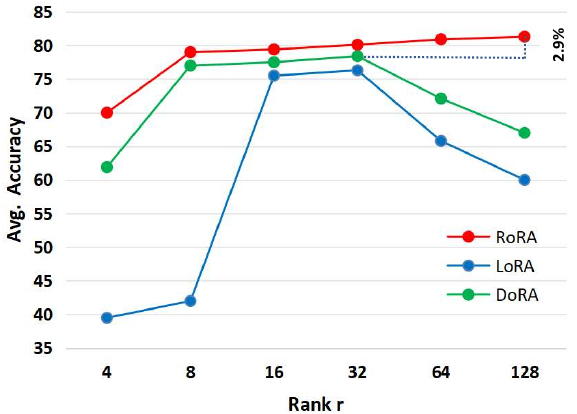}}
\caption{Average accuracy of LoRA, DoRA, and ours RoRA for varying ranks for LLaMA-7B on the commonsense reasoning tasks.}
\label{fig:comp}
\vskip 0in
\end{figure}
Our goal is to maximize fine-tuning performance under resource constraints. 
Fully fine-tuning is costly, and Low-Rank Adaptation (LoRA) \cite{hu2022lora} offers an efficient Parameter-Efficient Fine-Tuning (PEFT) approach for language and vision~\cite{dong2024physical},~\cite{dong2024df},~\cite{liu2021explainable},~\cite{zhan2024fast},~\cite{meng2025instructgie} models. The rank  \(r\) determines the dimensionality of low-rank weight updates, balancing resource efficiency and performance, with LoRA suggests that increasing the rank does not necessarily enhance subspace.
We observed that the performance of both LoRA and its improved SOTA version, Weight-Decomposed Low-Rank Adaptation (DoRA)~\cite{liu2024dora}, declines beyond \(r=32\), consuming more GPU without gains, as shown by the blue and green line~\cite{liu2024dora} in Fig.~\ref{fig:comp}.
%Our investigation reveals that this issue is influenced by the scaling factor used in LoRA, where gradient changes during fine-tuning are affected by rank size, leading to a decline in accuracy.

To address this problem, we propose a method that uses an optimized scaling factor (OpS) $\alpha/\sqrt{r}$ for fine-tuning both uncompressed and pruned~\cite{yuan2021work},~\cite{liu2024efficient},~\cite{yuan2022mobile},~\cite{li-etal-2020-efficient-transformer}\cite{shen2024search},\cite{zhan-etal-2024-rethinking-token}
%~\cite{liu2024efficient},~\cite{yuan2021work},~\cite{yuan2022mobile} 
LLMs. While rsLoRA~\cite{kalajdzievski2023rank} employs a similar scaling factor to study the
impact of the scaling factor on the learning process, our motivation, theoretical derivation, and experimental design are independent. This scaling factor mitigates the impact of rank, ensuring that gradient updates remain independent of rank. Our proposed method, Reliability Optimization for Rank Adaptation (RoRA), outperforms both LoRA and DoRA in fine-tuning uncompressed and pruned models.
The main contributions of our work are summarized as follows:
\begin{itemize}
\item LoRA's performance improvement was limited and declined with increasing rank. A mathematical analysis identified the scaling factor as a key factor in this decline.
% To address this, we proposed the RoRA method.
\item We propose RoRA to address gradient instability caused by varying ranks using the optimization scaling factor (OpS) $\alpha/\sqrt{r}$. 
This approach ensures that gradient changes are independent of rank, enhancing stability and performance during optimization. 
\item Extensive experiments on the commonsense reasoning dataset (see Fig.~\ref{fig:comp}) show that RoRA outperforms LoRA and DoRA in average accuracy by 6.5\% and 2.9\% on LLaMA-7B, respectively.
\item RoRA method consistently improved performance with increasing rank, marking its first use in fine-tuning both pruned and uncompressed large models.
\end{itemize}

\section{Background and Problem Formulation}
Pre-trained models like LLaMA ~\cite{touvron2023llama} possess broad linguistic knowledge but may lack domain~\cite{liu2022efficient},~\cite{liu2023interpretable},~\cite{liu2024brain} specialization. Users often fine-tune them with task-specific data to enhance performance while maintaining overall language.

LoRA~\cite{hu2022lora} is an efficient and widely used fine-tuning method. In LoRA, the update of \(m_0\) is constrained by a low-rank decomposition: \(m_0 + \Delta m = m_0 + \gamma B A\), where \( B\in \mathbb{R}^{p_{\text{out}} \times r}\), \( A\in \mathbb{R}^{r \times p_{\text{in}}}\), and \(\gamma\) is the scaling factor. During training, \(m_0\) remains fixed and does not receive gradient updates, while \(B\) and \(A\) are trainable parameters. The forward pass is given by:
\begin{equation}\label{eq:flora}
\mathcal{H}(x) = m_0x + \Delta mx = (m_0 + \gamma B A) x, 
\end{equation} 
where $x \in \mathbb{R}^{p_{\text{in}}}$ is the input, $\mathcal{H}(x) \in \mathbb{R}^{p_{\text{out}}}$ is the output, $m_0 \in \mathbb{R}^{p_{\text{out}} \times p_{\text{in}}}$.
The scaling factor $\gamma$ is set to $\alpha/r$ in ~\Cref{eq:flora}. 

\section{The Proposed Method}
To address the accuracy drop with increasing LoRA rank $r$, we analyzed the relationship between gradient variance and rank $r$. We optimized the scaling factor from $\alpha/r$ to $\alpha/\sqrt{r}$ to ensure that the gradient variance remains unaffected by rank $r$. This optimized scaling factor is defined as OpS (Optimization Scaling) and the method is referred to as Reliability Optimization for Rank Adaptation (RoRA).
\label{sec:method}

\subsection{Mathematics Analysis for the Weight Variance}  
The relationship between the gradient and its variance is critical: the gradient indicates the loss function's rate of change, while the variance reflects stability. We analyze how rank 
{r} affects both using mathematical techniques.

\noindent{\textbf{Mathematics Analysis.}}
We represent the increment part of the output $\mathcal{H}$ in~\cref{eq:flora} by $\mathbf{w} \in \mathbb{R}^{p_{\text{out}}}$,  $\mathbf x \in \mathbb{R}^{p_{\text{in}}}$ is the input, and replace the scaling factor $\alpha/r$ with $\gamma$. Each term $w_i$ can be expressed as:

\begin{equation}
\mathbf{w}_i = \gamma \sum_{j=1}^{p_{\text{in}}} \sum_{k=1}^r \mathbf{B}_{ik}\mathbf{A}_{kj}x_j. \label{eq:w_i_2}
\end{equation}

Using the chain rule to compute the partial derivatives of the loss function \( L \) with respect to \( \mathbf{B}_{ik} \) and \( \mathbf{A}_{kj} \), we have
\(\frac{\partial L}{\partial \mathbf{A}_{kj}} = \gamma \sum_{i=1}^{p_{\text{out}}} \frac{\partial L}{\partial \mathbf{w}_i} \mathbf{B}_{ik} x_j\), and
\(\frac{\partial L}{\partial \mathbf{B}_{ik}} = \gamma \frac{\partial L}{\partial \mathbf{w}_i} \sum_{j=1}^{p_{\text{in}}} \mathbf{A}_{kj} x_j\).
LoRA~\cite{hu2022lora} sets the learning rate \(\eta\) and initializes \(\mathbf{B}_{ik}=0\)~\cite{7410480,mishkin2015all}, a common optimization assumption to analyze early training and parameter scaling effects.
After the first step update, \(\mathbf{A}_{kj}\) remains unchanged, and \(\mathbf{B}_{ik}\) is updated as:
\begin{equation}
\mathbf{B}_{ik}^{(t+1)} = \mathbf{B}_{ik}^{(t)} - \eta \frac{\partial L}{\partial \mathbf{B}_{ik}} = -\eta \frac{\partial L}{\partial \mathbf{w}_i} \gamma \sum_{j=1}^{p_{\text{in}}} \mathbf{A}_{kj}x_j^{(t)}, \label{eq:bik2}
\end{equation}
where $\mathbf{B}_{ik}^{(t)}$ represents the before updated value of $\mathbf{B}_{ik}$, and $\mathbf{B}_{ik}^{(t+1)}$ represents its updated value after the previous step.

Substituting~\Cref{eq:bik2} into~\Cref{eq:w_i_2}, and replacing $\partial L/\partial \mathbf{w}_i$ with $\delta_i$, we have:
\begin{equation}
\mathbf{w}_i^{(t+1)}=-\eta \delta_i\gamma^2\sum_{j=1}^{p_{in}}\sum_{k=1}^r\sum_{l=1}^{p_{in}}\mathbf{A}_{kl}x_l^{(t)}\mathbf{A}_{kj}x_j^{(t+1)}.
\end{equation}

Assuming that $\delta_i$ is bounded and independent of $r$, and that the elements in $A$ and the inputs $x^{(t)}$ and $x^{(t+1)}$ are independently and identically distributed normal variables with mean 0 and variance 1, the variance of $\mathbf{w}_i^{(t+1)}$ is:

\begin{equation}
\text{Var}[\mathbf{w}_i^{(t+1)}] = \text{E}[(\mathbf{w}_i^{(t+1)})^2] - (\text{E}[\mathbf{w}_i^{(t+1)}])^2.
\end{equation}

Since the elements in the matrix $\mathbf{A}$ and the two inputs $x^{(t)}$ and $x^{(t+1)}$ are assumed to be independently, the expected value of $\mathbf{w}_i^{(t+1)}$ is :

\begin{equation}
\text{E}[\mathbf{w}_i^{(t+1)}] = -\eta \delta_i \gamma^2 \sum_{j=1}^{p_{\text{in}}} \sum_{k=1}^r \sum_{l=1}^{p_{\text{in}}} \text{E}[\mathbf{A}_{kl}] \text{E}[x_l^{(t)}] \text{E}[\mathbf{A}_{kj}] \text{E}[x_j^{(t+1)}].
\end{equation}

Since the expected values of $A$ and $x$ are both zero, the expected value of $\mathbf{w}_i^{(t+1)}$ is 0.
The variance is equal to the expected value squared: 

\begin{equation}
\text{Var}[\mathbf{w}_i^{(t+1)}] = \text{E}[{(\mathbf{w}_i^{(t+1)})}^2]\ .
\end{equation}

Therefore, we need to compute $\text{E}[{(\mathbf{w}_i^{(t+1)})}^2]$.
Substituting the expression for $\mathbf{w}_i^{(t+1)}$ and using the linearity of expectations have:
\begin{equation}
\begin{split}
\text{E}[{(\mathbf{w}_i^{(t+1)})}^2] &= \eta^2 \delta_i^2 \gamma^4 \sum_{j=1}^{p_{\text{in}}} \sum_{k=1}^r \sum_{l=1}^{p_{\text{in}}} \\
&\quad \text{E}[{(\mathbf{A}_{kl})}^2] \text{E}[{(x_l^{(t)})}^2] \text{E}[{(\mathbf{A}_{kj})}^2] \text{E}[{(x_j^{(t+1)})}^2].
\end{split}
\end{equation}

Since the variance of $A$ and $x$ is 1, we have:$\text{E}[\mathbf{A}_{kl}^2] = \text{E}[\mathbf{A}_{kj}^2] = 1$, $\text{E}[(x_l^{(t)})^2] = \text{E}[(x_j^{(t+1)})^2] = 1$,
thus, we  derived the expression:
\begin{equation}
\text{Var}[\mathbf{w}_i^{(t+1)}] = \text{E}[{(\mathbf{w}_i^{(t+1)})}^2] = \eta^2 \delta_i^2 \gamma^4 r^2 p_{\text{in}}.
\end{equation}

Now, to compute the magnitude of the magnitude of $\|\mathbf{w}\|_2$, we take the square root of the sum of the squares of all $\mathbf{w}_i^{(t+1)}$, which gives:

\begin{figure}[tbp]
\centerline{\includegraphics[width=0.99\linewidth]{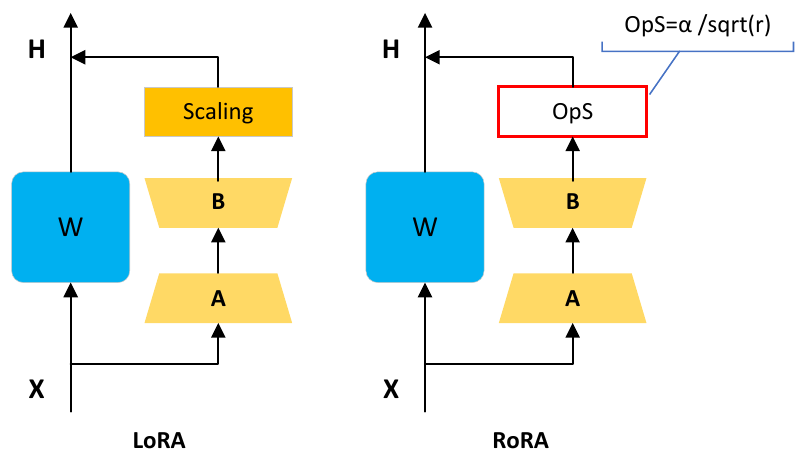}}
\caption{Difference between LoRA and RoRA.}
\label{fig:diff}
\vskip -0in
\end{figure}

\begin{equation}
\|\mathbf{w}\|_2 = \sqrt{\sum_{i=1}^{p_{\text{out}}} \left( \eta \delta_i \gamma^2 \sum_{j=1}^{p_{\text{in}}} \sum_{k=1}^r \sum_{l=1}^{p_{\text{in}}} \mathbf{A}_{kl} x_l^{(t)} \mathbf{A}_{kj} x_j^{(t+1)} \right)^2},
\end{equation}
where $\|\mathbf{w}\|_2$ is composed of many terms, each with coefficients $\eta \delta_i \gamma^2$ and products of matrix elements. The sum of the squares of these terms is approximately the maximum term value multiplied by the number of terms. Thus, $\|\mathbf{w}\|_2$ can be approximated by $\sqrt{p_{\text{out}}}$ times the maximum term value, where the maximum value is $\eta^2 \gamma^4 r^2 p_{\text{in}} \max(\delta_i^2)$ for $i \in (1, p_{\text{out}})$.

\begin{equation}
\|\mathbf{w}\|_2 \approx \sqrt{p_{\text {out}} \cdot \eta^{2} \gamma^{4} r^{2} p_{\text {in}} \cdot \max_{\substack{i \in (1, p_{\text{out}})}} \delta_{i}^{2}}.
\end{equation}

When $\max_{i \in \left(1, p_{\text{out}}\right)} \delta_i $ is denoted as $\delta_{m}$, $\|\mathbf{w}\|_{2}$ can be expressed as:
\begin{equation}
\begin{aligned}
  \|\mathbf{w}\|_2& \approx {c} \cdot \gamma^2 \cdot  {r},
 \end{aligned}
\end{equation}
where $\sqrt{p_{\text{out}}p_{\text{in}}}\eta \delta_{m}$ is represented by a constant $c$, as rank $r$ increases, the complexity of the function is approximately a certain function of $r$, this function does not exceed a constant multiple of $r$. $\mathcal{O}_r(r)$ can be used to indicate the complexity, $\|\mathbf{w}\|_{2}$ can be expressed as:
\begin{equation}
\|\mathbf{w}\|_2 \approx {c} \cdot \gamma^2 \cdot \mathcal{O}_r(r).
\end{equation}

\subsection{Reliability Optimization for Rank Adaptation}
By substituting the scaling factor $\gamma$ $=$ $\alpha/r$, which suggested by~\cite{hu2022lora},  we can obtain:
\begin{equation}
\|\mathbf{w}\|_2 \approx {c} \cdot \alpha^2 \cdot \mathcal{O}_r(1/r), 
\end{equation}
where $\mathcal{O}_r(1/r)$ means that as $r$ increases, its magnitude is approximately around $1/r$. The formula shows that the training output increment  $\mathcal{H}$ in~\Cref{eq:flora} slows down as rank $r$ increases.
If replace $\alpha/r$ with $\alpha/\sqrt{r}$, The following can be obtained:
\begin{equation}
\|\mathbf{w}\|_2 \approx {c} \cdot \alpha^2 \cdot \mathcal{O}_r(1), 
\end{equation}
where $\mathcal{O}_r(1)$ means the growth rate or complexity of the function is approximately constant as $r$ changes under given conditions. Therefore, it can be concluded that if the scaling factor is $\alpha/\sqrt{r}$, the change in the gradient is not affected by the rank. When the rank is small, the scaling factor is large; when the rank is large, the scaling factor is small. This adjustment helps to balance the impact of different ranks and provides better control over the gradient magnitude. 

We define the optimization scaling factor $\alpha/\sqrt{r}$ as OpS (Optimization Scaling). This method, named RoRA (Reliability Optimization for Rank Adaptation), uses OpS to effectively optimize the scaling factor.

\subsection{Comparison with LoRA}

In Fig.~\ref{fig:diff}, RoRA distinguishes itself from LoRA by thoroughly examining gradient variance. Our analysis shows that the rank \(r\) in LoRA can cause gradient instability. Using the optimization scaling factor (OpS) \(\alpha/\sqrt{r}\), we ensure that rank does not affect gradient changes, effectively mitigating this problem.

\section{EXPERIMENTS}
The performance of RoRA was evaluated on LLaMA-7B/13B~\cite{touvron2023llama}, LLaMA2-7B, LLaMA3-8B~\cite{llama3modelcard}, and the pruned model SHEARED-LLAMA-1.3B~\cite{xia2023sheared} on commonsense reasoning tasks using one NVIDIA A6000 48G GPU. RoRA was compared with LoRA~\cite{hu2022lora}, DoRA~\cite{liu2024dora}, and various baselines, including Prompt Learning (Prefix)\cite{li2021prefixtuning}, Series Adapter\cite{houlsby2019series}, and Parallel Adapter~\cite{he2022parallel}. The commonsense reasoning tasks included BoolQ~\cite{clark-etal-2019-boolq}, PIQA~\cite{Bisk2020piqa}, SIQA~\cite{sap-etal-2019-social}, HellaSwag~\cite{zellers2019hellaswag},WinoGrande~\cite{ai2:winogrande}, ARC-e~\cite{allenai:arc}, ARC-c~\cite{allenai:arc}, and OBQA~\cite{OpenBookQA2018}.

\begin{figure}[tbp]
\centerline{\includegraphics[width=0.78\linewidth]{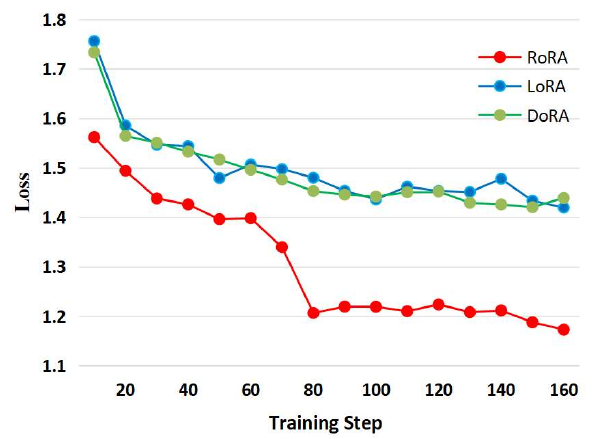}}
\caption{Comparison of the loss curves of LoRA, DoRA, and RoRA fine-tuning LLaMA 7B with $rank$ $r$ of 128.}
\label{fig:loss}
%\vskip -0.1in
\end{figure}

\begin{table*}[t]
\caption{Accuracy comparison of LLaMA 7B/13B, LLaMA2 7B, and LLaMA3 8B with various PEFT~\cite{peft} methods on eight commonsense reasoning datasets. Results for all methods are obtained using the hyperparameters described in DoRA~\cite{liu2024dora }}
%\vskip -0.05in
%\vskip 0.1in
\setlength{\tabcolsep}{1.2mm}
\centering
% \renewcommand\arraystretch{0.8}
%\resizebox{0.90\textwidth}{!}{
 % Adjust column spacing
%\setlength{\tabcolsep}{2.6pt} % Default is usually around 6pt, reduce to save space
\begin{tabular}{lc|ccccccccc|c}
\toprule
\textbf{Model} & \textbf{PEFT Method} & \# \textbf{Params (\%)} & \textbf{BoolQ} & \textbf{PIQA}&\textbf{SIQA}& \textbf{HellaS} & \textbf{WinoG}& \textbf{ARC-e} & \textbf{ARC-c} & \textbf{OBQA} & \textbf{Average$\uparrow$} \\ \hline

\multirow{6}{*}{LLaMA-7B}    & Prefix & 0.11 & 64.3 & 76.8 & 73.9 & 42.1 & 72.1 & 72.9 & 54.0 & 60.6 & 64.6 \\ 
    & Series & 0.99 & 63.0 & 79.2 & 76.3 & 67.9 & 75.7 & 74.5 & 57.1 & 72.4 & 70.8 \\ 
 & Parallel & 3.54 & 67.9 & 76.4 & 78.8 & 69.8 & 78.9 & 73.7 & 57.3 & 75.2 & 72.2 \\ 
                                & LoRA & 0.83 & 68.9 & 80.7 & 77.4 & 78.1 & 78.8 & 77.8 & 61.3 & 74.8 & 74.7 \\ 
                                 & DoRA & 0.84 & 69.7 &	83.4 &	78.6 &	87.2 &	81.0 &	81.9 &	66.2 &	79.2 &	78.4 \\
                                 & \ours & 0.83& 70.8 &82.3 &86.9 &87.7 &82.8 &82.9 &78.6	&78.6	 &\textbf{81.3} \\ \hline
\multirow{6}{*}{LLaMA-13B}     &Prefix & 0.03 & 65.3 & 75.4 & 72.1 & 55.2 & 68.6 & 79.5 & 62.9 & 68.0 & 68.4 \\ 
    & Series & 0.80 & 71.8 & 83 & 79.2 & 88.1 & 82.4 & 82.5 & 67.3 & 81.8 & 79.5 \\ 
 & Parallel & 2.89 & 72.5 & 84.9 & 79.8 & 92.1 & 84.7 & 84.2 & 71.2 & 82.4 & 81.4 \\ 
                                & LoRA & 0.67 & 72.1 & 83.5 & 80.5 & 90.5 & 83.7 & 82.8 & 68.3 & 82.4 & 80.5 \\ 
                                
                                 &DoRA & 0.68 & 72.4 & 84.9 & 81.5 & 92.4 & 84.2 & 84.2 & 69.6 & 82.8 & {81.5} \\ 
                                 & \ours& 0.67 &72.8	&86.3	&82.1	&95.2	&86.3	&88.4	&77.3	&87.6	&\bf84.5\\ \hline
\multirow{3}{*}{LLaMA2-7B}    & LoRA & 0.83 & 69.8&	79.9&	79.5&	83.6&	82.6&	79.8&	64.7&	81.0&	77.6 \\ 

    & DoRA & 0.84 & 71.8 &	83.7&	76.0&	89.1&	82.6&	83.7&	68.2&	82.4&	79.7 \\
    & RoRA & 0.84 & 75.4 &  87.3&   79.5&   92.7&   86.1&   87.4&   71.7&   85.3&
	
    \textbf{83.2} \\
    \hline
\multirow{3}{*}{LLaMA3-8B}    & LoRA & 0.70 &70.8&	85.2&	79.9&	91.7&	84.3&	84.2&	71.2&	79.0&	80.8 \\ 

    & DoRA & 0.71 & 74.6&	89.3&	79.9&	95.5&	85.6&	90.5&	80.4&	85.8& 85.2\\	
    & RoRA & 0.71 &76.6	&91.7	&82.2	&97.4	&87.6	&92.5	&82.3	&87.8 &	
    \textbf{87.3} \\ \hline %\bottomrule
\end{tabular}
\label{tab:llama_commonsense}
%\vskip -0.0in
\end{table*}

\begin{table*}[h]
    \centering
    
    \caption{Accuracy comparison of Sheared-LLaMA-1.3B~\cite{xia2023sheared} (81.4\% Pruned from LLaMA2-7B) with various PEFT~\cite{peft} methods on eight commonsense reasoning datasets. % In each section's first row represents the experimental results for LoRA~\cite{hu2022lora}, and the second row represents our experimental RoRA results. 
   Results for all methods are obtained using the hyperparameters described in DoRA~\cite{liu2024dora}.
    } \label{tbl:shared_rank}
    %\vskip -0.0in
    %\resizebox{0.85\linewidth}{!}{
    \begin{tabular}{lc|cccccccc|c}
        \toprule
       \bf Rank &\bf PEFT Method  &\bf BoolQ &\bf PIQA &\bf SIQA &\bf HellaS &\bf WinoG &\bf ARC-e &\bf ARC-c &\bf OBQA &\textbf{Average$\uparrow$} \\
        \midrule
        %\multirow{\parbox{1.6cm}{Rank=8 }}  
        \multirow{3}{*}{Rank=8}
        &LoRA  &50.5	&52.3	&26.0 	&41.5	&37.2	&27.5	&41.8	&43.2	&40.0 \\
        
        &DoRA &51.6	&53.2	&27.1	&42.2 	&39.8	&28.9	&42.8 	&44.9	&41.3\\ 

        &RoRA  &54.1 	&56.6	&31.0 	&53.5	&40.8	&29.4	&44.2 	&62.1	&46.5 \\
        \midrule
        %\multirow{\parbox{1.6cm}{Rank=16 }}  
        \multirow{3}{*}{Rank=16}
        &LoRA  &60.8	&61.7 	&32.2	&51.8	&51.5 	&36.8	&51.8 	&62.7 	&51.1\\
        
        &DoRA &61.9	&63.3	&34.3	&53.8	&52.9	&38.2	&53.6 	&64.0	&52.7\\ 

        &RoRA  &62.3	&62.0	&52.6	&52.9	&49.7	&36.8	&50.6	&64.0	&53.9 \\
        \midrule
       %\multirow{\parbox{1.6cm}{Rank=32 }} 
       \multirow{3}{*}{Rank=32}
      
       &LoRA(Best)  &60.4	&62.5	&37.0 	&53.0 	&49.5	&37.6	&48.2 	&61.7	&51.2 \\ 
       
       &DoRA(Best) &61.6	&64.6	&38.4	&54.5	&53.7	&38.1	&50.4	&64.3	&53.0 \\

       &RoRA  	&61.9	&63.3	&46.6	&53.8	&52.9	&38.9	&53.6 	&64.0	&54.4 \\
       \midrule
       %\multirow{\parbox{1.6cm}{Rank=64 }}  
       \multirow{3}{*}{Rank=64}
       &LoRA &60.3	&61.9 	&32.9	&52.3	&50.6 	&36.7	&51.1 	&63.1 	&51.1  \\  
       
       &DoRA &61.7	&63.3	&34.3	&53.8	&52.9	&38.2	&53.6 	&64.0	&52.7\\ 

       &RoRA &62.3 	&66.1	&46.6	&54.8	&53.6	&38.9	&53.6 	&64.3	&55.0 \\
       \midrule
       %\multirow{\parbox{1.6cm}{Rank=128 }}  
       \multirow{3}{*}{Rank=128}
       
       &LoRA &60.6	&67.0	&46.2	&53.1	&39.3	&27.3	&32.6 	&46.6 	&46.6  \\

       &DoRA &61.2	&67.6	&46.8	&53.2 	&39.8	&27.7	&32.8 	&47.0  &47.0\\
        &RoRA(\bf{Best})  	&62.7	&69.8 	&45.8	&58.6	&55.8	&40.6	&55.6 	&66.4	&\bf56.9 \\

    \bottomrule
    \end{tabular}
   %}
\vskip -0.05in  
\end{table*}

Table~\ref{tab:llama_commonsense} shows that~\ours~ average accuracy outperforms all baselines on LLaMA-7B/13B, LLaMA2-7B, and LLaMA3-8B. In commonsense reasoning tasks with ranks $r$ from 4 to 128, RoRA steadily improves, peaking at 81.3\% accuracy at rank 128, surpassing LoRA (74.7\%) and DoRA (78.4\%) by 6.5\% and 2.9\%, respectively (Fig.~\ref{fig:comp}). We also tested \(r=256\) on LLaMA-7B, LoRA and DoRA drop 2\%, while RoRA improves by 0.3\%, confirming its advantage as \(r\) approaches full fine-tuning with diminishing returns.

The loss curve in Fig.~\ref{fig:loss} illustrates the performance of three fine-tuning methods applied to LLaMA-7B, specifically with a rank of 128. RoRA shows a rapid initial drop in loss, followed by a steady decrease after step 60. Notably, RoRA achieves the lowest loss among all methods.
On a single NVIDIA A6000 GPU, LoRA and RoRA require about 3h 45m for Rank {r=8}, DoRA takes 5h 35m, and all methods add ~6m for Rank {r=128}, with inference latency about 29s.

Fine-tuning pruned models is more challenging than unpruned ones due to information loss and reduced flexibility, making hyperparameter tuning crucial. Comparing RoRA, DoRA, and LoRA in fine-tuning pruned models highlights their effectiveness in addressing these challenges.  
ShearedLLaMA~\cite{xia2023sheared} 1.3B is a pruned version of LLaMA2-7B, with an 81.4\% pruning rate, reducing it to 1.3 billion parameters. We fine-tuned it using RoRA, DoRA, and LoRA. Table 2 shows that LoRA and DoRA perform best at rank 32, while RoRA peaks at rank 128, achieving 3.9\% higher performance than DoRA and 5.7\% higher than LoRA at their optimal ranks. This demonstrates RoRA's significant advantage in fine-tuning pruned models compared to LoRA and DoRA.
\vskip 0.2in 
\section{CONCLUSION}
We introduce RoRA (Rank-adaptive Reliability Optimization), a sample yet effective method for optimizing the scaling factor in LoRA. By substituting $\alpha/r$ with $\alpha/\sqrt{r}$, RoRA improves performance as rank size increases, enhancing the subspace of low-rank adaptation matrices. This approach excels in fine-tuning both uncompressed and pruned models. Through extensive experiments, RoRA demonstrates effectiveness, achieving superior average accuracy and robustness compared to current state-of-the-art methods.
\vfill\pagebreak
\bibliographystyle{IEEEbib}
\bibliography{refs}

\begin{thebibliography}{10}

\bibitem{together2023redpajama}
Together Computer,
\newblock ``Redpajama-data: An open source recipe to reproduce llama training dataset,'' \url{https://github.com/togethercomputer/RedPajama-Data}, 2023.

\bibitem{puma3b}
Du~Bohan,
\newblock ``Openllama 3b v2 finetuned on sharegpt,'' \url{https://huggingface.co/acrastt/Puma-3B}, 2023.

\bibitem{Mamba-GPT-3b-v2}
chiliu,
\newblock ``Mamba-gpt-3b-v2,'' \url{https://huggingface.co/CobraMamba/mamba-gpt-3b-v2}, 2023.

\bibitem{openalpaca}
Yixuan Su, Tian Lan, and Deng Cai,
\newblock ``Openalpaca: A fully open-source instruction-following model based on openllama,'' \url{https://github.com/yxuansu/OpenAlpaca}, 2023.

\bibitem{xia2023sheared}
Mengzhou Xia, Tianyu Gao, et~al.,
\newblock ``Sheared llama: Accelerating language model pre-training via structured pruning,''
\newblock {\em arXiv preprint arXiv:2310.06694}, 2023.

\bibitem{zhang2022advancing}
Yihua Zhang, Yuguang Yao, Parikshit Ram, et~al.,
\newblock ``Advancing model pruning via bi-level optimization,''
\newblock {\em NeurIPS}, vol. 35, pp. 18309--18326, 2022.

\bibitem{li2022pruning}
Yanyu Li, Pu~Zhao, et~al.,
\newblock ``Pruning-as-search: Efficient neural architecture search via channel pruning and structural reparameterization,''
\newblock {\em International Joint Conference on Artificial Intelligence (IJCAI-22)}, 2022.

\bibitem{yang2023pruning}
Changdi Yang, Pu~Zhao, Yanyu Li, et~al.,
\newblock ``Pruning parameterization with bi-level optimization for efficient semantic segmentation on the edge,''
\newblock in {\em CVPR}, 2023, pp. 15402--15412.

\bibitem{yuan2022you}
Geng Yuan, Sung-En Chang, et~al.,
\newblock ``You already have it: A generator-free low-precision dnn training framework using stochastic rounding,''
\newblock in {\em ECCV}. Springer, 2022, pp. 34--51.

\bibitem{yuan2022layer}
Geng Yuan, Yanyu Li, et~al.,
\newblock ``Layer freezing \& data sieving: missing pieces of a generic framework for sparse training,''
\newblock {\em NeurIPS}, vol. 35, pp. 19061--19074, 2022.

\bibitem{yuan2021mest}
Geng Yuan, Xiaolong Ma, et~al.,
\newblock ``Mest: Accurate and fast memory-economic sparse training framework on the edge,''
\newblock {\em NeurIPS}, vol. 34, pp. 20838--20850, 2021.

\bibitem{li2024waxing}
Sheng Li et~al.,
\newblock ``Waxing-and-waning: a generic similarity-based framework for efficient self-supervised learning,''
\newblock in {\em ICLR}, 2024.

\bibitem{zhan2024exploring}
Zheng Zhan, Zhenglun Kong, Yifan Gong, et~al.,
\newblock ``Exploring token pruning in vision state space models,''
\newblock in {\em The Conference on Neural Information Processing Systems}, 2024.

\bibitem{zhao-etal-2024-pruning}
Pu~Zhao, Fei Sun, et~al.,
\newblock ``Pruning foundation models for high accuracy without retraining,''
\newblock in {\em Findings of the Association for Computational Linguistics: EMNLP 2024}, 2024.

\bibitem{yuan2021forms}
Geng Yuan, Payman Behnam, et~al.,
\newblock ``Forms: Fine-grained polarized reram-based in-situ computation for mixed-signal dnn accelerator,''
\newblock in {\em 2021 ACM/IEEE 48th Annual International Symposium on Computer Architecture (ISCA)}. IEEE, 2021, pp. 265--278.

\bibitem{yuan2021tinyadc}
Geng Yuan, Payman Behnam, et~al.,
\newblock ``Tinyadc: Peripheral circuit-aware weight pruning framework for mixed-signal dnn accelerators,''
\newblock in {\em 2021 Design, Automation \& Test in Europe Conference \& Exhibition (DATE)}. IEEE, 2021, pp. 926--931.

\bibitem{gong2022automatic}
Yifan Gong, Geng Yuan, et~al.,
\newblock ``Automatic mapping of the best-suited dnn pruning schemes for real-time mobile acceleration,''
\newblock {\em ACM Transactions on Design Automation of Electronic Systems (TODAES)}, vol. 27, no. 5, pp. 1--26, 2022.

\bibitem{peft}
Sourab Mangrulkar, Sylvain Gugger, et~al.,
\newblock ``Peft: State-of-the-art parameter-efficient fine-tuning methods,'' \url{https://github.com/huggingface/peft}, 2022.

\bibitem{liu2024tsla}
Jun Liu et~al.,
\newblock ``Tsla: A task-specific learning adaptation for semantic segmentation on autonomous vehicles platform,''
\newblock {\em IEEE Transactions on Computer-Aided Design of Integrated Circuits and Systems}, 2024.

\bibitem{liu2023scalable}
Jun Liu et~al.,
\newblock ``A scalable real-time semantic segmentation network for autonomous driving,''
\newblock in {\em Advanced Multimedia Computing for Smart Manufacturing and Engineering (AMC-SME)}, 2023, pp. 3--12.

\bibitem{hu2022lora}
Edward~J Hu, yelong shen, et~al.,
\newblock ``Lo{RA}: Low-rank adaptation of large language models,''
\newblock in {\em ICLR}, 2022.

\bibitem{dong2024physical}
Haoye Dong, Tiange Xiang, Sravan Chittupalli, Jun Liu, and Dong Huang,
\newblock ``Physical-space multi-body mesh detection achieved by local alignment and global dense learning,''
\newblock in {\em WACV}, 2024, pp. 1267--1276.

\bibitem{dong2024df}
Haoye Dong, Jun Liu, and Dong Huang,
\newblock ``Df-vton: Dense flow guided virtual try-on network,''
\newblock in {\em ICASSP}, 2024, pp. 3175--3179.

\bibitem{liu2021explainable}
Jun Liu, Feng Deng, et~al.,
\newblock ``An explainable convolutional neural networks for automatic segmentation of the left ventricle in cardiac mri.,''
\newblock in {\em CECNet}, 2021, pp. 306--314.

\bibitem{zhan2024fast}
Zheng Zhan, Yushu Wu, et~al.,
\newblock ``Fast and memory-efficient video diffusion using streamlined inference,''
\newblock in {\em Conference on Neural Information Processing Systems}, 2024.

\bibitem{meng2025instructgie}
Zichong Meng, Changdi Yang, Jun Liu, Hao Tang, Pu~Zhao, and Yanzhi Wang,
\newblock ``Instructgie: Towards generalizable image editing,''
\newblock in {\em European Conference on Computer Vision}. Springer, 2025, pp. 18--34.

\bibitem{liu2024dora}
Shih-Yang Liu, Chien-Yi Wang, et~al.,
\newblock ``Dora: Weight-decomposed low-rank adaptation,''
\newblock {\em arXiv preprint arXiv:2402.09353}, 2024.

\bibitem{yuan2021work}
Geng Yuan, Peiyan Dong, et~al.,
\newblock ``Work in progress: Mobile or fpga? a comprehensive evaluation on energy efficiency and a unified optimization framework,''
\newblock in {\em 2021 IEEE 27th Real-Time and Embedded Technology and Applications Symposium (RTAS)}, 2021, pp. 493--496.

\bibitem{liu2024efficient}
Jun Liu, Zhenglun Kong, et~al.,
\newblock ``Efficient pruning of large language model with adaptive estimation fusion,''
\newblock {\em arXiv preprint arXiv:2403.10799}, 2024.

\bibitem{yuan2022mobile}
Geng Yuan, Peiyan Dong, et~al.,
\newblock ``Mobile or fpga? a comprehensive evaluation on energy efficiency and a unified optimization framework,''
\newblock {\em ACM Transactions on Embedded Computing Systems}, vol. 21, no. 5, pp. 1--22, 2022.

\bibitem{li-etal-2020-efficient-transformer}
Bingbing Li, Zhenglun Kong, et~al.,
\newblock ``Efficient transformer-based large scale language representations using hardware-friendly block structured pruning,''
\newblock in {\em Findings of the Association for Computational Linguistics: EMNLP 2020}, Nov. 2020, pp. 3187--3199.

\bibitem{shen2024search}
Xuan Shen, Pu~Zhao, Yifan Gong, Zhenglun Kong, et~al.,
\newblock ``Search for efficient large language models,''
\newblock in {\em Advances in Neural Information Processing Systems}, 2024.

\bibitem{zhan-etal-2024-rethinking-token}
Zheng Zhan, Yushu Wu, Zhenglun Kong, et~al.,
\newblock ``Rethinking token reduction for state space models,''
\newblock in {\em the 2024 Conference on Empirical Methods in Natural Language Processin.}, 2024.

\bibitem{kalajdzievski2023rank}
Damjan Kalajdzievski,
\newblock ``A rank stabilization scaling factor for fine-tuning with lora,''
\newblock {\em arXiv preprint arXiv:2312.03732}, 2023.

\bibitem{touvron2023llama}
Hugo Touvron, Thibaut Lavril, et~al.,
\newblock ``Llama: Open and efficient foundation language models,''
\newblock {\em arXiv preprint arXiv:2302.13971}, 2023.

\bibitem{liu2022efficient}
Jun Liu, Feng Deng, et~al.,
\newblock ``An efficient cnn for radiogenomic classification of low-grade gliomas on mri in a small dataset,''
\newblock {\em Wireless Communications and Mobile Computing}, vol. 2022, no. 1, 2022.

\bibitem{liu2023interpretable}
Jun Liu, Geng Yuan, et~al.,
\newblock ``An interpretable cnn for the segmentation of the left ventricle in cardiac mri by real-time visualization.,''
\newblock {\em CMES-Computer Modeling in Engineering \& Sciences}, vol. 135, no. 2, 2023.

\bibitem{liu2024brain}
Jun Liu, Geng Yuan, et~al.,
\newblock ``Brain tumor classification on mri in light of molecular markers,''
\newblock {\em arXiv preprint arXiv:2409.19583}, 2024.

\bibitem{7410480}
Kaiming He, Xiangyu Zhang, et~al.,
\newblock ``Delving deep into rectifiers: Surpassing human-level performance on imagenet classification,''
\newblock in {\em ICCV}, 2015, pp. 1026--1034.

\bibitem{mishkin2015all}
Dmytro Mishkin and Jiri Matas,
\newblock ``All you need is a good init,''
\newblock {\em arXiv preprint arXiv:1511.06422}, 2015.

\bibitem{llama3modelcard}
AI@Meta,
\newblock ``Llama 3 model card,'' \url{https://github.com/meta-llama/llama3/blob/main/MODEL_CARD.md}, 2024.

\bibitem{li2021prefixtuning}
Xiang~Lisa Li and Percy Liang,
\newblock ``Prefix-tuning: Optimizing continuous prompts for generation,''
\newblock in {\em ACL-IJCNLP (Volume 1: Long Papers)}, 2021, pp. 4582--4597.

\bibitem{houlsby2019series}
Neil Houlsby, Andrei Giurgiu, et~al.,
\newblock ``Parameter-efficient transfer learning for nlp,''
\newblock in {\em ICML}, 2019, pp. 2790--2799.

\bibitem{he2022parallel}
Junxian He, Chunting Zhou, et~al.,
\newblock ``Towards a unified view of parameter-efficient transfer learning,''
\newblock in {\em ICLR}, 2021.

\bibitem{clark-etal-2019-boolq}
Christopher Clark, Kenton Lee, et~al.,
\newblock ``{B}ool{Q}: Exploring the surprising difficulty of natural yes/no questions,''
\newblock in {\em NAACL: Human Language Technologies, Volume 1}, 2019, pp. 2924--2936.

\bibitem{Bisk2020piqa}
Yonatan Bisk, Rowan Zellers, et~al.,
\newblock ``Piqa: Reasoning about physical commonsense in natural language,''
\newblock in {\em AAAI}, 2020.

\bibitem{sap-etal-2019-social}
Maarten Sap, Hannah Rashkin, et~al.,
\newblock ``Social {IQ}a: Commonsense reasoning about social interactions,''
\newblock in {\em EMNLP-IJCNLP}, Nov. 2019, pp. 4463--4473.

\bibitem{zellers2019hellaswag}
Rowan Zellers, Ari Holtzman, et~al.,
\newblock ``Hellaswag: Can a machine really finish your sentence?,''
\newblock in {\em ACL}, 2019.

\bibitem{ai2:winogrande}
Keisuke Sakaguchi, Ronan~Le Bras, et~al.,
\newblock ``Winogrande: An adversarial winograd schema challenge at scale,'' 2019.

\bibitem{allenai:arc}
Peter Clark, Isaac Cowhey, et~al.,
\newblock ``Think you have solved question answering? try arc, the ai2 reasoning challenge,''
\newblock {\em arXiv:1803.05457v1}, 2018.

\bibitem{OpenBookQA2018}
Todor Mihaylov, Peter Clark, et~al.,
\newblock ``Can a suit of armor conduct electricity? a new dataset for open book question answering,''
\newblock in {\em EMNLP}, 2018.

\end{thebibliography}

\end{document}